\documentclass{llncs}
\usepackage[table]{xcolor}
\usepackage{amsmath}
\usepackage{graphicx}        
\usepackage{multicol}        
\usepackage{multirow}
\usepackage{lscape}
\usepackage{float}
\usepackage{url}
\usepackage{caption,subfig}
\usepackage[utf8]{inputenc}
\usepackage{color}

\def\dae{{\em Divide-and-Evolve}}
\def\DAE{{\sc DaE}}
\def\DAEX{{\sc DaE$_{\text{X}}$}}
\newcommand{\DAEYAHSP}{{\sc DaE$_{\text{YAHSP}}$}}
\newcommand{\DAECPT}{{\sc DaE$_{\text{CPT}}$}}
\def\PARADISEO{{\sc ParadisEO-MOEO}}
\def\YAHSP{{\sc YAHSP}}
\def\CPT{{\sc CPT}}

\def\ZENO{{\sc Zeno}}
\def\MULTIZENO{{\sc MultiZeno}}
\def\PARAMILS{{\sc ParamILS}}

\def\MODAEYAHSP{{\sc MO-DaE$_{\text{YAHSP}}$}}
\def\AGGDAEYAHSP{{\sc Agg-DaE$_{\text{YAHSP}}$}}

\def\WMAKESPAN{{W-makespan}}
\def\WCOST{{W-cost}}

\begin{document}

\mainmatter              
\title{Multi-Objective AI Planning: Comparing Aggregation and Pareto Approaches}
%
%
 \author{M.~R. Khouadjia \inst{1} \and M. Schoenauer\inst{1}\and
 V. Vidal\inst{2}  \and J. Dr\'eo\inst{3} \and P. Sav\'eant\inst{3}}

%
%
 \institute{TAO Project, INRIA Saclay \&  LRI Paris-Sud University, Orsay, France\\
 \email{\{mostepha-redouane.khouadjia, marc.schoenauer\}@inria.fr},\\ 
 \and
 ONERA-DCSD, Toulouse, France\\
 \email{Vincent.Vidal@onera.fr}\\
 \and
 THALES Research \& Technology, Palaiseau, France\footnote{This work is being partially funded by the French National Research Agency under the research contract DESCARWIN (ANR-09-COSI-002)}\\
 \email{\{johann.dreo, pierre.saveant\}@thalesgroup.com}\\
}

\maketitle              

\begin{abstract}
Most real-world Planning problems are multi-objective, trying to minimize both the makespan of the solution plan, and some cost of the actions involved in the plan. But most, if not all existing approaches are based on single-objective planners, and use an aggregation of the objectives to remain in the single-objective context. \dae\ is an evolutionary planner that won the temporal deterministic satisficing track at the last International Planning Competitions (IPC). Like all Evolutionary Algorithms (EA), it can easily be turned into a Pareto-based Multi-Objective EA. It is however important to  validate the resulting algorithm by comparing it with the aggregation approach: this is the goal of this paper. The comparative experiments on a recently proposed benchmark set that are reported here demonstrate the usefulness of going Pareto-based in AI Planning.
\end{abstract}
\section{Introduction}

Most, if not all, classical AI planning solvers are single-objective. Given a planning domain (a set of predicates that describe the state of the system, and a set of actions with their pre-requisites and effects), and an instance of this domain (a set of objects on which the predicates are instantiated into boolean atoms, an initial state and a goal state), classical planners try to find, among the set of all feasible plans (sequences of actions such that, when applied to the initial state, the goal state becomes true), the one with the minimal number of actions (STRIP planning), or with the smallest cost (actions with costs) or with the smallest makespan (temporal planning, where actions have durations and can be applied in parallel). A detailed introduction to (single-objective) AI planning can be found in \cite{AIplanningBook2004}. It is clear, however, that most planning problems are in fact multi-objective, as the optimal solution in real-world problems often involve some trade-off between makespan and 
cost \cite{kambhampati-invited-icaps03}. A few trials have been made to turn some classical planners into multi-objective optimizers, either using some twist in PDDL 2.0\footnote{Planning Domain Definition Language, a dedicated language for the description of planning problems, set up for the International Planning Competitions (IPC).} to account for both makespan and cost \cite{do2003sapa,refanidis2003multiobjective,gerevini2008}, or using the new hooks for several objectives offered by PDDL 3.0 \cite{gerevini2006preferences}. However, all these approaches are based on a linear aggregation of the different objectives, and were not pursued, as witnessed by the new ``net-benefit'' IPC track, dedicated to aggregated multiple objectives, that took place in 2006 \cite{chen2006temporal} and 2008 \cite{edelkamp2009optimal}, \ldots but was canceled in 2011 due to a lack of entries. 

In the framework of Evolutionary Algorithms (EAs), Pareto multi-objective optimization has received a lot of attention \cite{Deb-book}, and any single-objective EA can ``easily'' be turned into a multi-objective EA, by modifying the selection step (and possible adding some archiving mechanism). Unfortunately, there exist very few evolutionary AI planners. Directly evolving plans, as in \cite{Morignot-2005}, obviously does not scale up, and was never extended to multi-objective setting. Hence, as far as we are aware of, the state-of-the-art in evolutionary AI planning is the previous work of some of the authors, \dae\ (\DAE). \DAE\ evolves variables length sequences of states, that start with the problem initial state and end at the problem goal state. \DAE\ relies on a classical embedded planner to sequentially reach each state of the sequence from the previous one. The concatenation of all plans given by the embedded planner is a solution plan of the original problem. \DAE\ can thus solve all types of 
planning problems that the embedded planner can solve. Proof-of-concept for \DAE\ was obtained with \DAECPT\ \cite{evoCOP2006}, where the embedded planner was \CPT, an exact planner \cite{vidal:aaai04} -- and already included some small multi-objective experiments. Since then, the \DAE\ paradigm has evolved, and \YAHSP\, a sub-optimal lookahead strategy planning system \cite{Vidal2004} is now used as the embedded planner \cite{bibai-EvoCOP2010}, and \DAEYAHSP\ has reached state-of-the-art results in all planning domains \cite{Bibai2010}, winning the temporal deterministic satisficing track at the last IPC in 2011\footnote{See {\tt http://www.plg.inf.uc3m.es/ipc2011-deterministic}}. 

The very preliminary work in \cite{evoCOP2006} regarding multi-objective optimization has also been recently revisited with \DAEYAHSP. The lack of existing benchmark suite for multi-objective planning led us to extend the small toy problem from \cite{evoCOP2006} into a tunable benchmark domain, on which different multi-objectivization of \DAEYAHSP\ (\MODAEYAHSP) were compared \cite{nous-emo2013}. But because the only other approach in AI Planning is the aggregation of the objectives, there is a need to compare the multi-objective approach for \DAEYAHSP\ with the single-objective approach based on the linear aggregation of the objectives: this is the purpose of the present work.
Section \ref{sec:DAE} will briefly present planning problems and \DAEYAHSP\ in the single-objective setting. In Section \ref{sec:multi}, the multi-objective context will be introduced. The multi-objective benchmark suite will be presented, and the multi-objectivization of \DAEYAHSP\ will be detailed: because \YAHSP\ is a single-objective planner\footnote{Note that it seems difficult, if at all opssible, to adapt it directly to multi-objective optimization, as it uses very different strategies for the makespan and the cost.}, but can be asked to optimize either the makespan or the cost, specific strategies had to be designed regarding how it is called within \MODAEYAHSP. Section \ref{sec:settings} describes the experimental settings, detailing in particular the implementation of the aggregation approach for \DAEYAHSP\ and the intensive parameter tuning that was performed for all competing algorithms using the off-line problem-independent tuner \PARAMILS\ \cite{ParamILS-JAIR}. The results will be presented and 
discussed in Section \ref{sec:results}, and as usual, conclusion and hints about on-going and further work will be given in Section \ref{sec:conclusion}.

\section{Single-Objective Background}
\label{sec:DAE}
\paragraph{AI Planning Problems:} A planning domain $D$ is defined by a set of object types, a set of predicates, and a set of possible actions. An instance is defined by a set of objects of the domain types, an initial state, and a goal state. A predicate that is instantiated with objects is called an atom, and takes a boolean value. For a given instance, a state is defined by assigning values to all possible atoms. An action is defined by a set of {\em pre-conditions} (atoms) and a set of {\em effects} (changing some atom values): the action can be executed only if all pre-conditions are true in the current state, and after an action has been executed, the state is modified: the system enters a new state. The goal is to find a plan (sequence of actions) such that it leads from the initial state to the goal state, and minimizes either the number or costs of actions, or the makespan in the case of temporal planning where actions have durations and can be run in parallel.

\begin{figure}[tb]
\begin{tabular}{ccc}
\parbox{7cm}{
 \includegraphics[height=0.5\textwidth,width=4cm,angle=270]{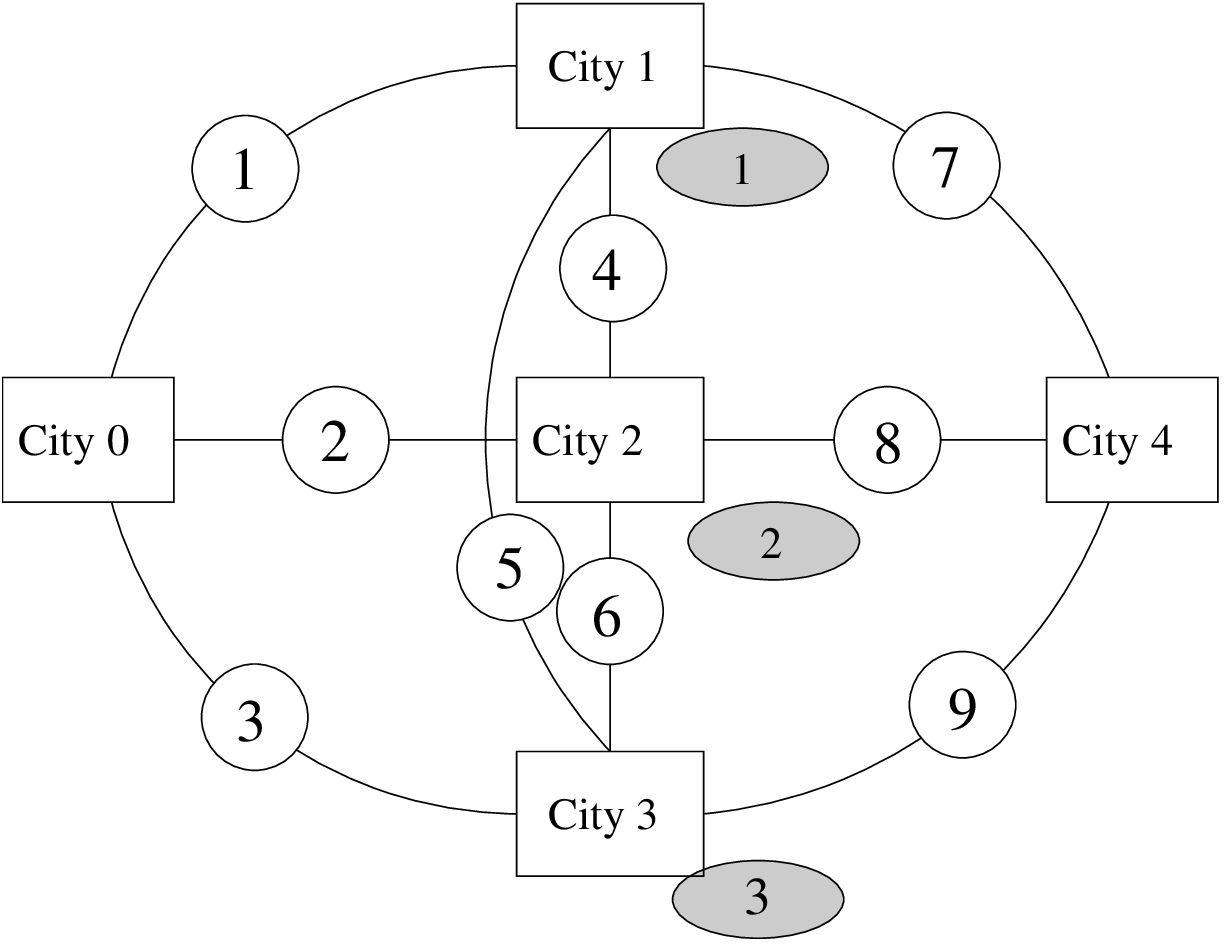} \\
{Flight durations are attached to the possible routes (white circles), costs/risks are attached to landing in the central cities (grey circles). Four sets of values given on the right. Default values in the first (``Lin.'') column in the Table.}
}
& ~~~ &
\begin{tabular}{|c||c|c|c|}
\hline
Dur./edge & Lin. & Cvx & Ccve  \\
\hline 
1 & 2 & 2 & 2\\
2 & 4 & 4 & {\bf 3}\\
3 & 6 & 6 & {\bf 4}\\
4 & 3 & 3 & {\bf 1}\\
5 & 5 & 5 & {\bf 2}\\
6 & 3 & 3 & {\bf 1}\\
7 & 2 & 2 & 2\\
8 & 4 & 4 & {\bf 3}\\
9 & 6 & 6 & {\bf 4}\\
\hline \hline
Cost/city & & & \\
\hline
1 &  30 & 30 & 30\\
2 & 20 & {\bf 11} &{\bf 29}\\
3 & 10 & 10 & 10 \\
\hline
\end{tabular}
\end{tabular}
\caption{Schematic view, and 3 instances, of simple \MULTIZENO\ benchmark.}
\vskip -0.4cm
\label{fig.instance}
\end{figure}

A simple temporal planning problem in the domain of logistics (inspired by the well-known {\ZENO} problem of IPC series) is given in Figure \ref{fig.instance}, and will be the basis of the benchmark used in this work: the problem involves cities, passengers, and planes (object types). Passengers can be transported from one city to another (action {\tt fly}), following the links on the figure. One plane can only carry one passenger at a time from one city to another, and the flight duration (number on the link) is the same whether or not the plane carries a passenger (this defines the {\em domain} of the problem). In the simplest non-trivial {\em instance} of such domain, there are 3 passengers and 2 planes. In the initial state, all passengers and planes are in {\tt city 0}, and in the goal state, all passengers must be in {\tt city 4}. In the default case labeled ``Lin.'' in the table right (forget about the costs for now), the not-so-obvious makespan-optimal solution has a total makespan of 8 and is left 
as a teaser for the reader.

\paragraph{Divide-and-Evolve:} Let ${\cal P}_D(I,G)$ denote the planning problem defined on domain $D$ with initial state $I$ and goal state $G$. In order to solve  ${\cal P}_D(I,G)$, the basic idea of \DAEX\ is to find a sequence of states $S_1, \ldots, S_n$, and to use some embedded planner $X$ to solve the series of planning problems ${\cal P}_D(S_{k},S_{k+1})$, for $k \in [0,n]$ (with the convention that $S_0 = I$ and $S_{n+1} = G$).
The generation and optimization of the sequence of states $(S_i)_{i \in [1,n]}$  is driven by an evolutionary algorithm. The fitness of a sequence is computed using the embedded planner $X$, that is called in turn on each of the sub-problems ${\cal P}_D(S_{k},S_{k+1})$. The concatenation of the corresponding plans (possibly compressed to take into account possible parallelism in the case of temporal planning) is a solution of the initial problem. In case one sub-problem cannot be solved by the embedded solver, the individual is said {\em unfeasible} and its fitness is highly penalized in order to ensure that unfeasible individuals are always selected after feasible ones. A thorough description of \DAEX\ can be found in \cite{Bibai2010}. 
The rest of this section will briefly recall the evolutionary parts of \DAEX.

An individual in \DAEX\ is a  variable-length list of partial states of the given domain (similar to the goal state), and a partial state is a variable-length list of atoms (instantiated predicates). The initialization procedure is based on a heuristic estimation, for each atom, of the earliest time from which it can become true~\cite{Haslum2000}. Furthermore, most existing planners (and this is true for \CPT\ and \YAHSP, that have been used within \DAE) start by computing some partial mutual exclusion between possible atoms: this information is also used to reduce the search space in \DAEX, whenever possible.
An individual in \DAEX\ is hence a variable-length time-consistent sequence of partial states, and each partial state is a variable-length list of atoms that are not pairwise mutually exclusive. 

Crossover and mutation operators are applied with respective user-defined probabilities $p_{Cross}$ and $p_{Mut}$. They are defined on the \DAEX\ representation in a straightforward manner - though constrained by the heuristic chronology and the partial mutex relation between atoms.
{\bf One-point crossover} is adapted to variable-length representation: both crossover points are independently chosen, uniformly in both parents. Only one offspring is kept, the one that respects the approximate chronological constraint on the successive states. 
{\bf Four different mutation operators} are included, and operate either at the individual level, by adding ({\tt addState}) or removing ({\tt delState}) a state, or at the state level by adding or modifying ({\tt addChangeAtom}) or removing ({\tt delAtom}) some atoms in a uniformly chose state. The choice among these operators is made according to user-defined relative weights (named w-mutationname - see Table \ref{tab:parameters}).

\section{Multi-Objective Background}
\label{sec:multi}
\subsection{Pareto-based Multi-Objective Divide-and-Evolve}
\label{sec:pareto-dae}
Two modifications of \DAEYAHSP\ are needed to turn it into an EMOA: use some multi-objective selection engine in lieu of the single-objective tournament selection that is used in the single-objective context; and compute the value of both objectives (makespan and cost) for both individuals. The former modification is straightforward, and several alternatives have been experimented within \cite{nous-emo2013}. The conclusion is that the indicator-based selection using the hypervolume difference indicator \cite{Zitzler2004} performs best -- and only this one will be used in the following, denoted here \MODAEYAHSP.
As explained above, the computation of the fitness is done by \YAHSP -- and \YAHSP, like all known planners to-date, is a single-objective planner. It is nevertheless possible, since PDDL 3.0 \cite{gerevini2006preferences}, to specify other quantities of interest that are to be computed throughout the execution of the final plan, without interfering with the search. Within \MODAEYAHSP, two strategies are then possible for \YAHSP: it can be asked to optimize either the makespan or the cost, and to simply compute the cost or the makespan when executing the solution plan (for feasible individuals). 

The choice between both strategies is governed by user-defined weights, named respectively \WMAKESPAN\ and \WCOST\ (see table \ref{tab:parameters}). For each individual, the actual strategy is randomly chosen according to those weights, and applied to all subproblems of the individual. Note that those weights are tuned using ParamILS (see Section \ref{sec:settings}), and it turned out that the optimal values for \MODAEYAHSP\ have always been equal weights: something that was to be expected, as no objective should be preferred to the other. 

\subsection{Aggregation-based Multi-Objective Divide-and-Evolve}
\label{sec:aggregation}
Aggregation is certainly the easiest and most common way to handle multi-objective problems with a single-objective optimization algorithm: a series of single-objective optimization problems are tackled in turn, the fitness of each of these problems is defined by a linear combination of the objectives. In the case of makespan and cost, both to be minimized, each linear combination can be defined by a single parameter $\alpha$ in $[0,1]$. In the following, $F_{\alpha}$ will denote $\alpha * \mbox{makespan} + (1-\alpha) * \mbox{cost}$, and \DAEYAHSP\ run optimizing $F_{\alpha}$ will be called the $\alpha$-run. One ``run'' of the aggregation method thus amounts to running several $\alpha$-runs, and returns the set of non-dominated individuals among the union of all final populations\footnote{Some adaptive method has been proposed \cite{adaptiveWeightsEMO01}, where parameter $\alpha$ is adapted on-line, spanning all values within a single run: this is left for further work.}. Note that different $\
alpha$-runs 
might have different optimal values for their parameters: a complete parameter tuning run of \PARAMILS\ must be performed for each $\alpha$-run to ensure a fair comparison with other well-tuned approaches.

The choice of the number of values to choose for the different $\alpha$ depends on the available resources. But the choice of the actual values aims at exploring the objective space as uniformly as possible, and some issues might arise if both objectives are not scaled similarly.
We hence propose here to use some evenly spaced values for $\alpha$ (see Section \ref{sec:settings}), but only after both objectives have been scaled into [0,1]. However, for such scaling to be possible, some bounds must be known for each objective. When they are not known, these bounds can be approximated from single-objective runs on each of the objectives in turn.

\begin{figure}[tb]
\centering{
\begin{tabular}{cccc}
\includegraphics[width=0.32\textwidth]{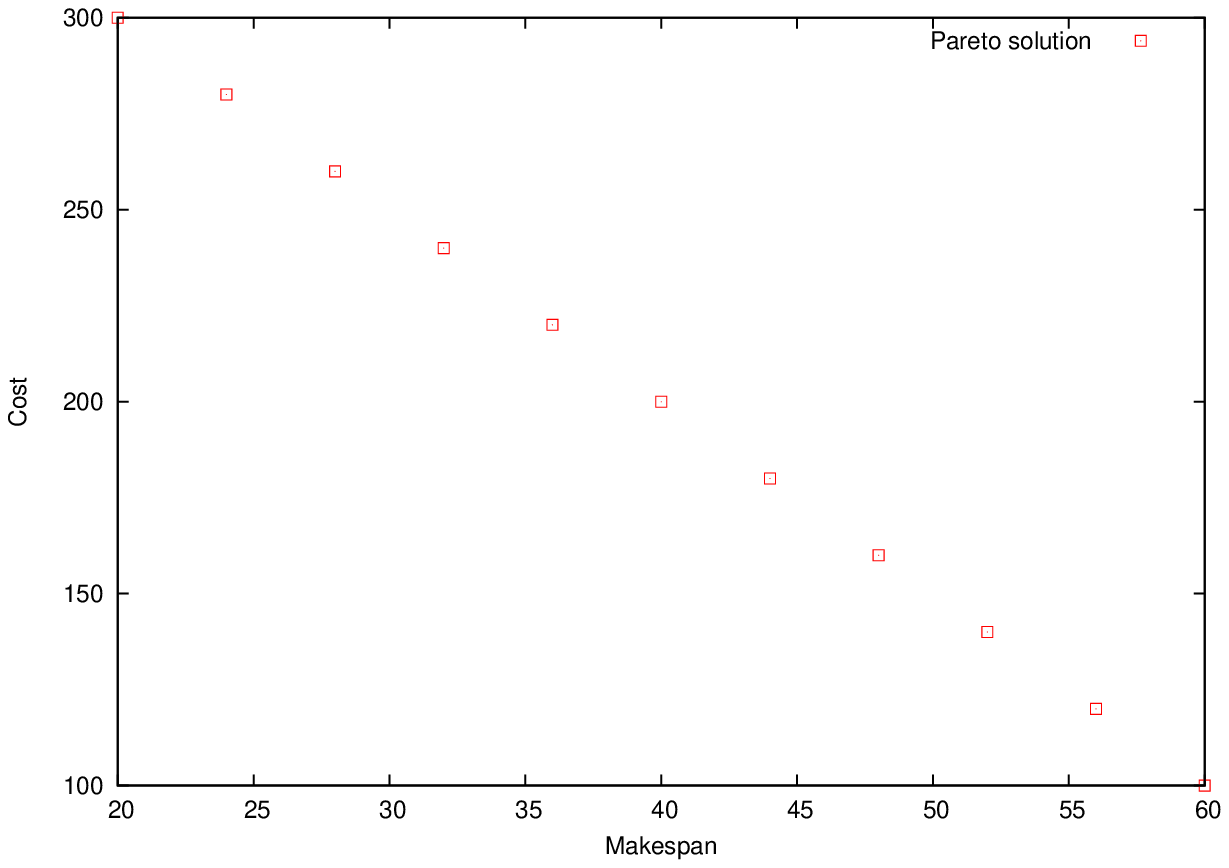} &
\includegraphics[width=0.32\textwidth]{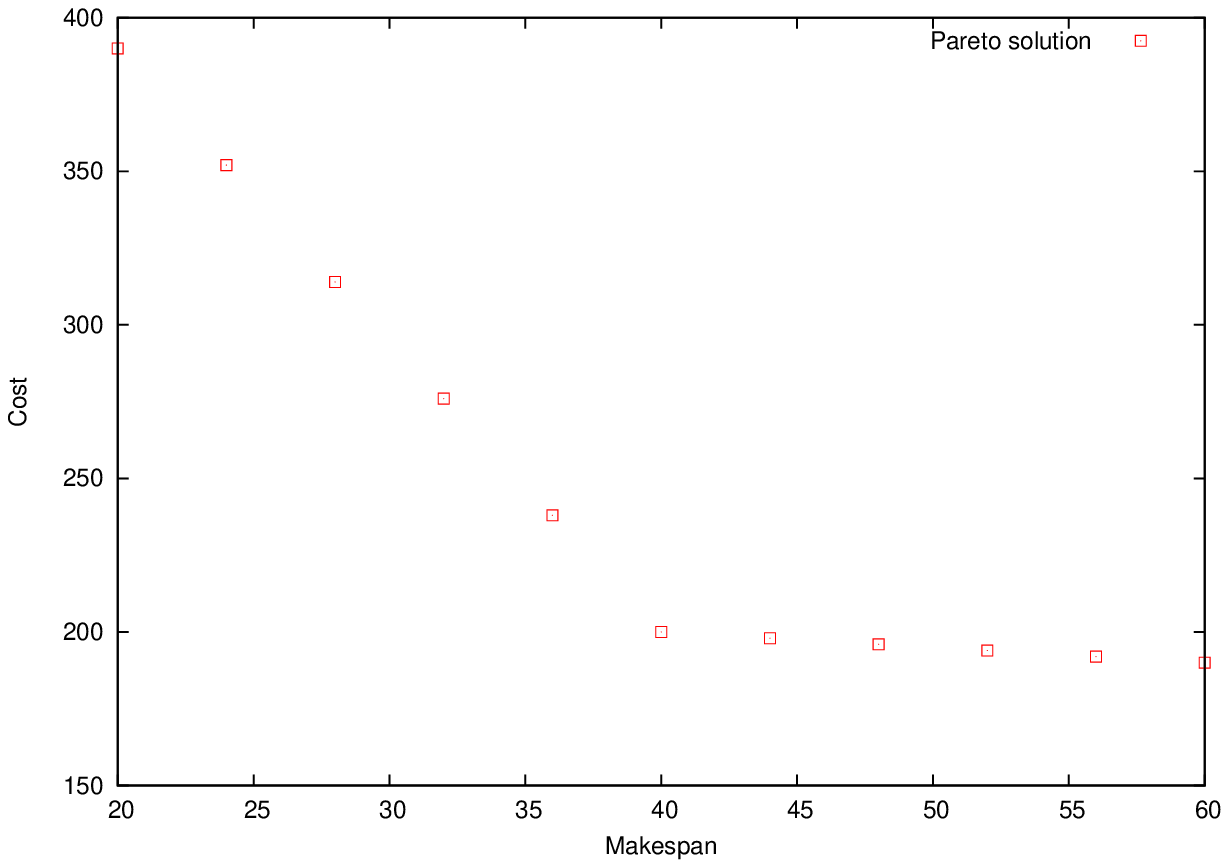} &
\includegraphics[width=0.32\textwidth]{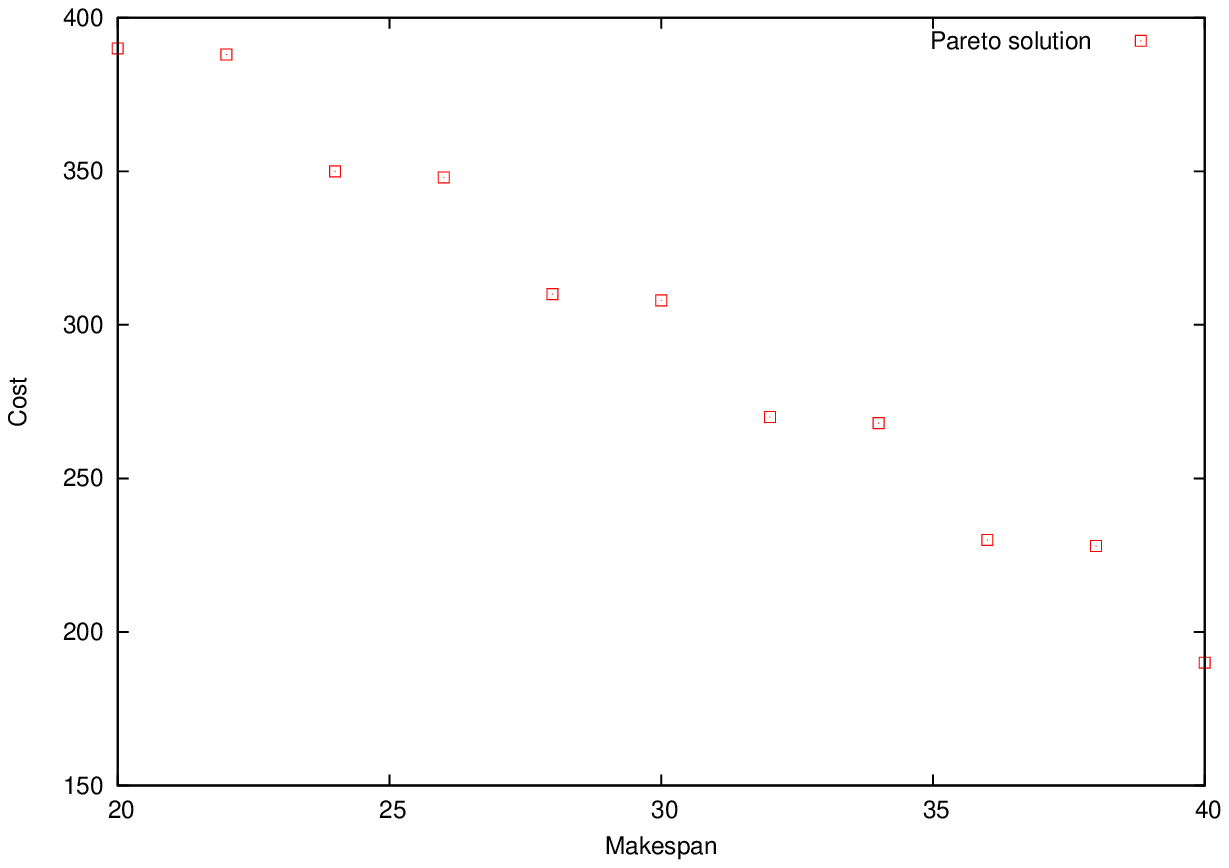}\\
Linear & Convex & Concave 
\end{tabular}
\caption{Pareto Fronts for the \MULTIZENO6$_{Cost}$ problems described in Figure \ref{fig.instance}.}
\label{fig:allParetoFronts}
}
\end{figure}

\subsection{Multi-Objective Benchmarks:}
The reader will have by now solved the little puzzle set in Section \ref{sec:DAE}, and found the solution with makespan 8, that manages to leave no plane idle (detailed solution in \cite{nous-emo2013}). In order to turn this problem into a multi-objective one, costs (or risks) are added to the {\tt fly} actions that land in one of the central cities, leading to two types of problem: In \MULTIZENO$_{Cost}$, the second objective is the total costs, that is accumulated every time a plane lands in a central city; In \MULTIZENO$_{Risk}$, the second objective is the maximal risk encountered during the complete execution of a plan; both are to be minimized. The complexity of the instances can be increased by adding more passengers: instances with 3, 6 and 9 passengers will be used here. Finally, by tuning the values of the flight durations and the costs/risks, different shapes of the Pareto front can be obtained: Figure \ref{fig.instance} summarizes three possible instances for the \MULTIZENO\ domain, and the 
corresponding Pareto fronts for the 6-passengers case are displayed in Figure \ref{fig:allParetoFronts}.

\section{Experimental Settings}
\label{sec:settings}

\begin{table}[tb!]
\scriptsize
\begin{tabular}{|l|l|l|}
\hline
Parameters 	&	   Range 	&	Description\\					
\hline									
\WMAKESPAN\	&	[0,5]	&	Weight for makespan strategy for \YAHSP \\
\WCOST\ 	&	[0,5]	&	  Weight for cost/risk strategy for \YAHSP \\	
\hline
Pop-size 	&	[10,300]	&	 Population size\\
\hline
Proba-cross	&	[0,1]	&	 Probability to apply cross-over\\	
Proba-mut	&	[0,1]	&	 Probability to apply one of the mutation\\	
\hline
w-addatom	&	[1,10]	&	 Weight for addChangeAtom mutation\\					
w-addgoal	&	[1,10]	&	 Weight for addGoal mutation\\					
w-delatom	&	[1,10]	&	 Weight for delAtom mutation\\					
w-delgoal	&	[1,10]	&	 Weight for delGoal mutation\\												
\hline
Proba-change	&	[0,1]	&	 Probability to change each atom in the addChangeAtom mutation\\							
Proba-delatom	&	[0,1]	&	 Probability to delete each atom in the delAtom mutation\\								
Radius	&	[1,10]	&	  Number of neighbour goals to consider for the addGoal mutation\\		
\hline				
\end{tabular}
\caption{Set of parameters off-line tuned using \PARAMILS.}
\label{tab:parameters}
\end{table}

\paragraph{Parameter Tuning:}
\label{sec:tuning}
It is now widely acknowledged that the large number of parameters of most EAs, even though it is a source of flexibility, is also a weakness, in that a poor parameter setting can ruin the performances of the most promising algorithm. Whereas no generic approach exists for on-line control, there are today many available methods for off-line parameter tuning that should be used within any evolutionary experiment, in spite of their huge computational cost. 

In this work, unless otherwise stated, the user-defined parameters of both \MODAEYAHSP\ and \DAEYAHSP\ shown in Table \ref{tab:parameters} have been tuned anew for each instance, using the \PARAMILS\ framework
\cite{ParamILS-JAIR}. \PARAMILS\ performs an Iterated Local Search in the space of possible parameter configurations, evaluating each configuration by running the algorithm to be optimized with this configuration on the given instance. 

\paragraph{Stopping Criteria:} Due to the variable number of calls to \YAHSP\, the number of function evaluation is not representative of the CPU effort of runs of \DAEYAHSP. Hence the stopping criterion of all \DAEYAHSP\ run was set to a given wall-clock time (300, 600 and 1800 seconds for \MULTIZENO3, 6 and 9 respectively (on an Intel(R) Xeon(R) @ 2.67GHz or equivalent). That of \MODAEYAHSP was set accordingly: for the sake of a fair comparison, because one run of the aggregated approach requires $n$ runs of the single-objective version of \DAEYAHSP, \MODAEYAHSP was run for $n$ times the time of each of the \DAEYAHSP\ runs. In the following, $n$ will vary from $3$ to $8$ (see Section \ref{sec:results}). The stopping criterion for \PARAMILS\ was likewise set to a fixed wall-clock time: 48h (resp. 72h) for \MULTIZENO3 and 6 (resp. \MULTIZENO9), corresponding to 576, 288, and 144 parameter configuration evaluations for \MULTIZENO3, 6 and 9 respectively.

\paragraph{Performance Metrics and Results Visualization:} The quality measure used by \PARAMILS\ to optimize the parameters of both \MODAEYAHSP\ and each of the $\alpha$-runs of \DAEYAHSP\ is the unary hypervolume  $I_{H^-}$~\cite{Zitzler2004} of the set of non-dominated points output by the algorithm with respect to the complete true Pareto front (only instances where the true Pareto front is fully known have been experimented with). The lower the better (a value of 0 indicates that the exact Pareto front has been reached). All reported differences in hypervolume have been tested using Wilcoxon signed rank test at 95\% confidence level, unless otherwise stated.

However, and because the true front is made of a few scattered points (at most 17 for \MULTIZENO9\ in this paper), it is also possible to visually monitor the empirical Cumulative Distribution Function of the probability to discover each point, as well as the whole front. This allows some deeper comparison between algorithms even when none has found the whole front. Such {\em hitting plots} will be used in the following, together with more classical plots of hypervolume vs time.
Finally, because hitting plots only tell if a given point was reached and do not provide any information regarding how far from the other points the different runs ended, more details on the approximated Pareto fronts will be given by visualizing the merged final populations of all runs of given settings.

\paragraph{Implementation:} For all experiments, 11 independent runs have been performed, implemented within the \PARADISEO\ framework\footnote{\url{http://paradiseo.gforge.inria.fr/}}. All performance assessment procedures (hypervolume calculations, statistical tests), have been achieved using the PISA performance assessment tool\footnote{\url{http://www.tik.ee.ethz.ch/pisa/}}.

\section{Experimental Results}
\label{sec:results}
This section will compare the Pareto-based \MODAEYAHSP\ and the aggregation approach \AGGDAEYAHSP\ on \MULTIZENO3, 6 and 9. Unless otherwise stated, the default domain definition leading to a linear Pareto front (see Figure \ref{fig.instance} and \ref{fig:allParetoFronts}-left) will be used, and one \AGGDAEYAHSP\ run will be made of 7 different $\alpha$-runs, with $\alpha$ taking the values $0, 0.1, 0.3, 0.5, 0.7, 0.9,$ and $1$.

\paragraph{The MultiZeno3 Problem} proved to be too easy: both \MODAEYAHSP\ and \AGGDAEYAHSP\ find the complete Pareto fronts, and the hitting plots reach 100\% in less than 80s (resp. 90s) for the {\sc Cost} (resp. {\sc Risk}) version of the instance (not shown here). \MODAEYAHSP\ is slightly slower (resp. faster) than \AGGDAEYAHSP\ in the {\sc Cost} (resp. {\sc Risk}) instance, but no significant difference is to be reported. Only instances -6 and -9 will be looked at in the following.

\paragraph{The Risk Objective:} On these instances, however, the {\sc Risk} objective proved to be almost too difficult to be of interest here, even though there are only 3 points on the Pareto Front, whatever the number of passengers: as can be seen on Figure \ref{fig:allZenosAttainments}, no algorithm could identify the complete Pareto front for the \MULTIZENO9\ instance (line 4); for \MULTIZENO6\ (line 2), \MODAEYAHSP\ could reliably identify the whole front (in 9 runs out of 11), while only  a single run of \AGGDAEYAHSP\ could identify the middle point (40,20). \MODAEYAHSP\ is hence a clear winner here - however, too little information is brought by the risk value, as one single stop in a risky station will completely hide the possibly low-risk remaining of the plan. Further work will aim at designing a smoother fitness for such situations.

The rest of the paper will hence concentrate on the {\sc Cost} versions of \MULTIZENO6 and 9 (simply denoted \MULTIZENO\{6,9\}), where significant differences between both approaches can be highlighted. 

\begin{figure}[tb]
\includegraphics[bb=50 50 410 302,width=0.48\textwidth,height=3cm]{zeno6x-hyperlogscale_{cost}.eps}
\includegraphics[bb=50 50 410 302,width=0.48\textwidth,height=3cm]{zeno9x-Add-hyper_{cost}.eps} 
\caption{Evolution of hypervolume for \DAEYAHSP\ (green squares) and \AGGDAEYAHSP\ (blue triangles) for \MULTIZENO6 (left) and \MULTIZENO9 (right).}
\label{fig:hypervolume}
\end{figure}

\begin{figure}[tb]
\includegraphics[bb=50 50 410 302,width=0.32\textwidth]{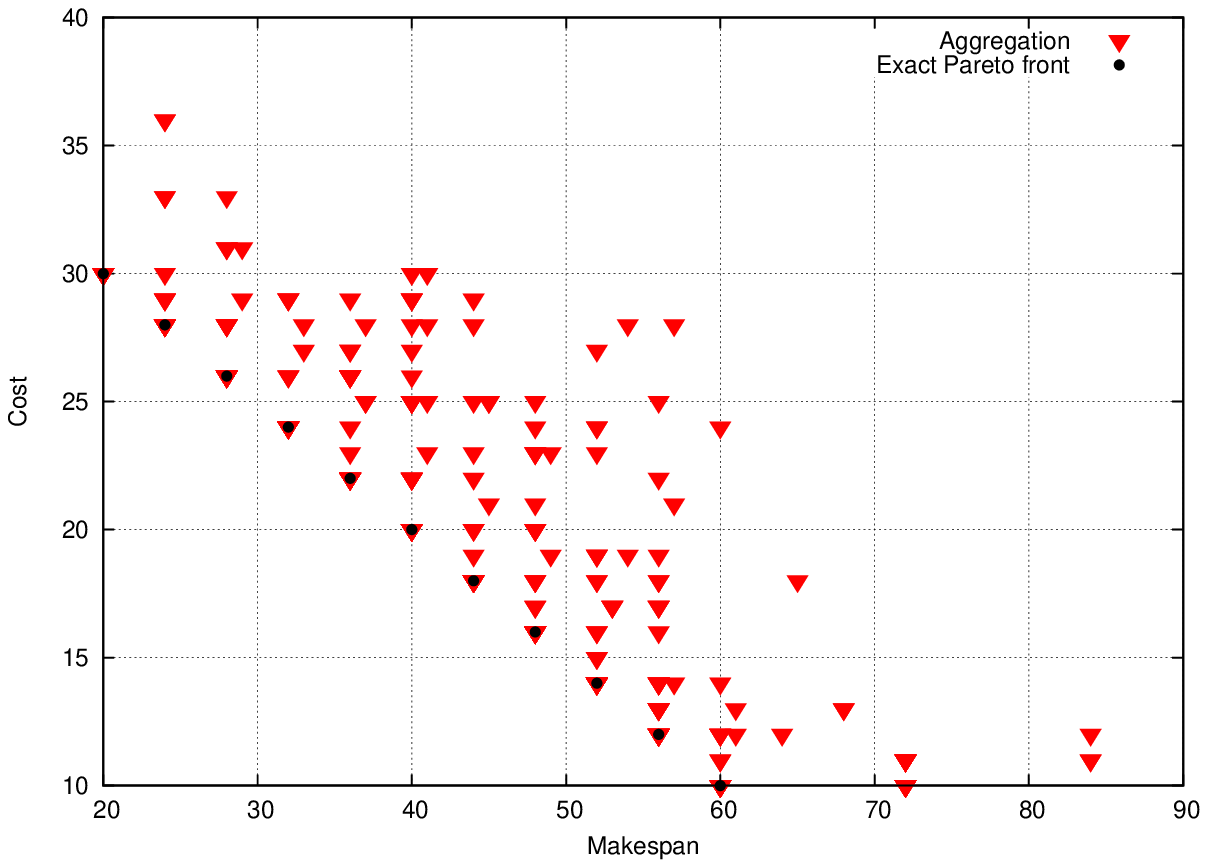}
\includegraphics[bb=50 50 410 302,width=0.32\textwidth]{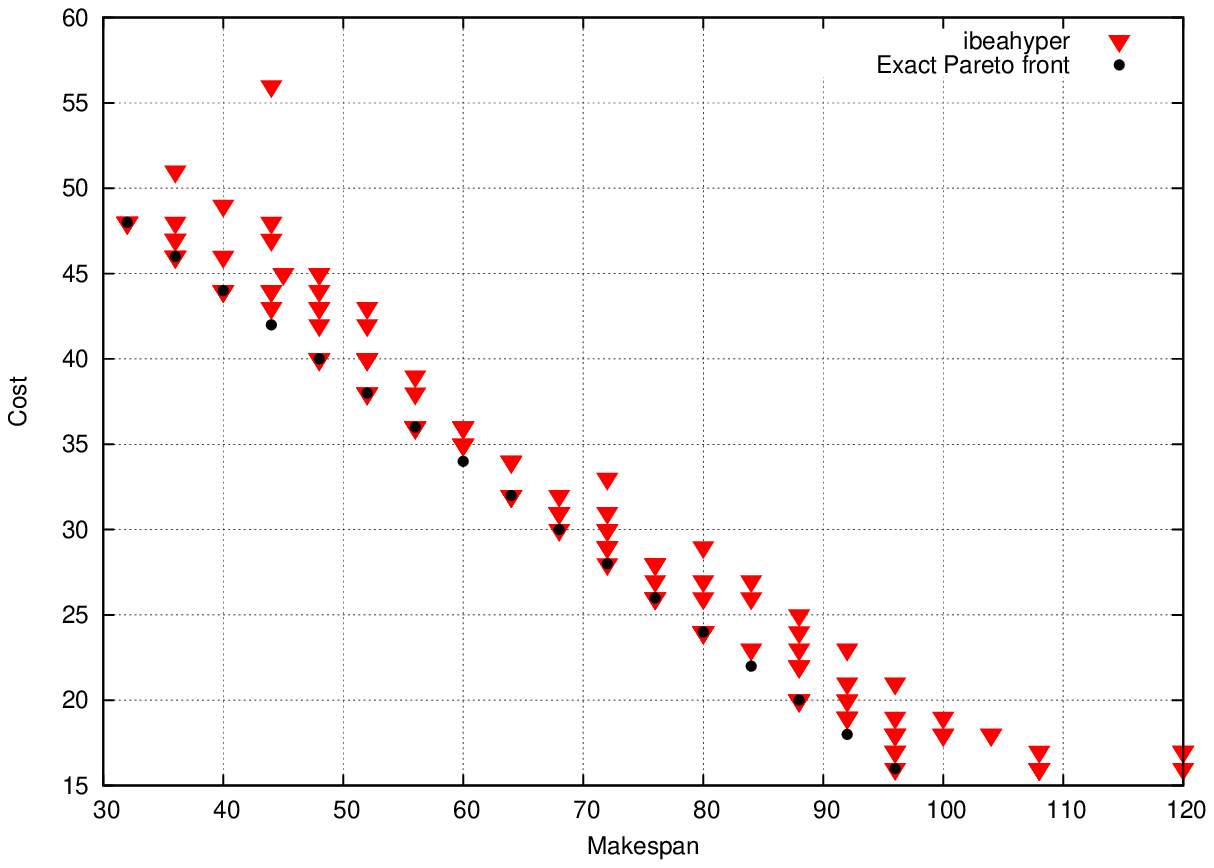}
\includegraphics[bb=50 50 410 302,width=0.32\textwidth]{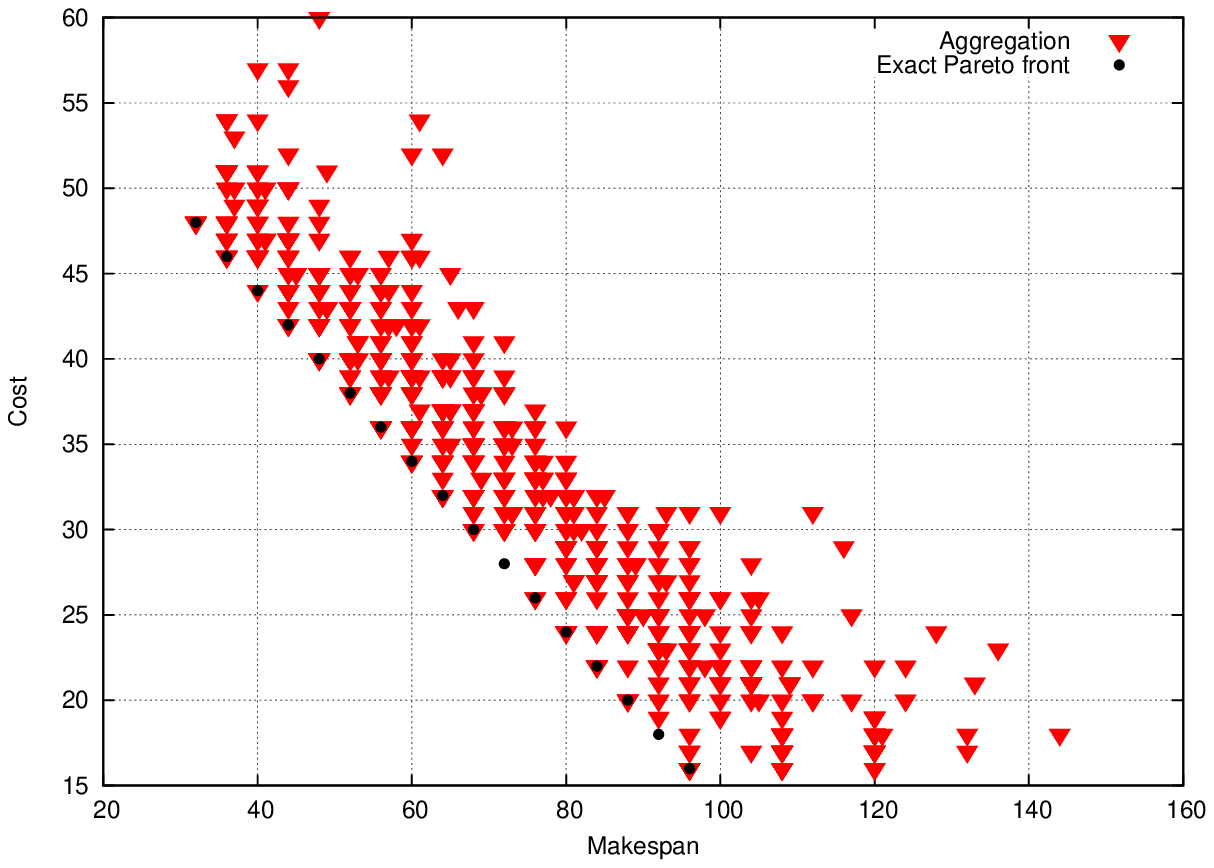}\\
~\parbox{0.30\textwidth}{\centering{\AGGDAEYAHSP\ on \MULTIZENO6}}~~~ 
\parbox{0.30\textwidth}{\centering{\MODAEYAHSP\ on \MULTIZENO9}}~~~
\parbox{0.30\textwidth}{\centering{\AGGDAEYAHSP\ on \MULTIZENO9}} 
\caption{Pareto fronts approximations (union of all final populations).}
\label{fig:approxPareto}
\vskip -0.5cm
\end{figure}

\paragraph{Results on the Default Instance:} From the plots of the evolution of the average hypervolumes (Figure \ref{fig:hypervolume}), \MODAEYAHSP\ is the winner for \MULTIZENO6, and \AGGDAEYAHSP\ is the winner for \MULTIZENO9. Taking a closer look at the hitting plots (Figure \ref{fig:allZenosAttainments}), we can see for \MULTIZENO6 (line 1) that all runs of \MODAEYAHSP\ reach the complete Pareto front in around 2500s, while only 9 runs out of 11 do reach it. On the other hand, for \MULTIZENO9, and though the figures of line 3 are more difficult to read because they contain the CDF for 17 points, slightly more points seem to be reached by \AGGDAEYAHSP\ than by \MODAEYAHSP. Looking now at the approximations of the Pareto fronts (Figure \ref{fig:approxPareto}), the fronts returned by \AGGDAEYAHSP\ for \MULTIZENO6 show a large dispersion away from the true front, whereas the same figure for \MODAEYAHSP\ (not shown) only contains the true front. Regarding \MULTIZENO9, even though it reaches less points 
from the true front, \MODAEYAHSP\ demonstrates a much more robust behavior than \AGGDAEYAHSP, for which the approximate fronts are, again, quite dispersed, sometimes far from the true front. 

\begin{figure}[tb]
\includegraphics[bb=50 50 410 302,width=0.48\textwidth,height=3.5cm]{zeno6dix_{cost}:IBEADED_{H^-}.eps} 
\includegraphics[bb=50 50 410 302,width=0.48\textwidth,height=3.5cm]{zeno6diDedalpha_{cost}:overallscale.eps}\\ 
\includegraphics[bb=50 50 410 302,width=0.48\textwidth,height=3.5cm]{zeno6sc_{cost}:IBEADED_{H^-}.eps}
\includegraphics[bb=50 50 410 302,width=0.48\textwidth,height=3.5cm]{zeno6scalphaded_{cost}:overallscale.eps}
\caption{Hitting plots for \MODAEYAHSP\ (left) and \AGGDAEYAHSP\ (right), for instances 2, 3, and 4 of \MULTIZENO6 from Figure \ref{fig.instance} (from top to bottom).}
\label{fig:convexConcave}
\vskip -0.5cm
\end{figure}

\paragraph{Results on other MultiZeno6 instances:} Further experiments have been conducted on different variants of \MULTIZENO6 instance, described in Figure \ref{fig.instance}. The corresponding hitting plots can be seen on Figure \ref{fig:convexConcave}. As in the {\sc Linear} default case, \MODAEYAHSP\ is a clear winner -- and this is confirmed by the plots of the approximate Pareto fronts (not shown), for which \AGGDAEYAHSP\ again shows a much larger dispersion away from the true front than \MODAEYAHSP. 

All results presented until now have been obtained by first optimizing the parameters of all algorithms with \PARAMILS. Interestingly, when using the parameters optimized by \PARAMILS\ for the Linear instance on these other instances, the results are only slightly worse: this observation will motivate further work dedicated to the generalization of the parameter tuning across instances.

\begin{figure}[htbp]
\includegraphics[bb=50 50 410 302,width=0.48\textwidth]{zeno6ex_{cost}:IBEA_{H^-}.eps}  
\includegraphics[bb=50 50 410 302,width=0.48\textwidth]{zeno6exalpha_{cost}:overallscale.eps}\\
\includegraphics[bb=50 50 410 302,width=0.48\textwidth]{zeno6ex_{risk}:IBEA_{H^-}.eps}
\includegraphics[bb=50 50 410 302,width=0.48\textwidth]{zeno6eMaxalpha_{risk}:overallscale.eps}\\
\includegraphics[bb=50 50 410 302,width=0.48\textwidth]{zeno9ex_{cost}:IBEA_{H^-}.eps}
\includegraphics[bb=50 50 410 302,width=0.48\textwidth]{zeno9exalpha_{cost}:overallscale.eps}\\
\includegraphics[bb=50 50 410 302,width=0.48\textwidth]{zeno9ex_{risk}:IBEA_{H^-}.eps}
\includegraphics[bb=50 50 410 302,width=0.48\textwidth]{zeno9eMaxalpha_{risk}:overallscale.eps}
\caption{Hitting plots for \MODAEYAHSP\ (left) and \AGGDAEYAHSP\ (right), for instances (from top to bottom) \MULTIZENO6$_{Cost}$, \MULTIZENO6$_{Risk}$, \MULTIZENO9$_{Cost}$, and \MULTIZENO9$_{Risk}$. The lower line on each plots is the experimental CDF for the probability to reach the whole Pareto front.}
\label{fig:allZenosAttainments}
\end{figure}

\section{Discussion and Conclusion}
\label{sec:conclusion}

\vskip -0.4cm
The experiments presented in this paper have somehow demonstrated the greater efficiency of the Pareto-based approach to multi-objective AI Planning \MODAEYAHSP\ compared to the more traditional approach by aggregation of the objectives \AGGDAEYAHSP. The case is clear on \MULTIZENO6, and on the different instances that have been experimented with, where \MODAEYAHSP\ robustly finds the whole Pareto front (except for the {\sc Convex} instance), whereas \AGGDAEYAHSP\ performs much worse in all aspects. 
This is also true on the \MULTIZENO9 instance, in spite of the better hypervolume indicator: indeed, a few more points on the Pareto front are found a little more often, but the global picture remains a poor approximation of the Pareto front. Other experiments on more instances are needed to confirm these first results, and on-going work is concerned with solving instances generated from IPC benchmarks by merging the cost and the temporal 
domains when the same instances exist in both.

Regarding the computational cost, one \AGGDAEYAHSP\ run requires several single-objective runs -- and as many parameter tuning procedures. We have chosen here to use 7 different values for $\alpha$, and it was clear from results not shown here that taking away a few of these resulted in a decrease of quality of the results. The computational cost of the parameter tuning could be reduced, too: first, a complete tuning anew for each instance is unrealistic, and was only done here for the sake of a fair comparison between both approaches; second, even on a single instance, it should be possible to  tune all parameters (except those of \YAHSP\ strategy) for all $\alpha$-runs together.
Finally, one of the most promising directions for future research is the on-line tuning of \YAHSP\ strategy, e.g., using a self-adaptive approach, where the strategies are attached to the individual.

{\small
\bibliographystyle{splncs}
\bibliography{khouadjia}
}

\end{document}